\documentclass[10pt,twocolumn]{ICCAS}
 
\usepackage{diagbox}

\usepackage{amsmath}
\usepackage{amssymb}
\usepackage{algorithm}
\usepackage{algpseudocode}
\usepackage[utf8x]{inputenc}
\usepackage{graphicx}
\usepackage{multirow}
\begin{document}

\title{VLM-driven Skill Selection for Robotic Assembly Tasks}

\author{Jeong-Jung Kim$^{1*}$, Doo-Yeol Koh$^{1}$ and Chang-Hyun Kim$^{1}$}

\affils{ ${}^{1}$Department of AI Machinery, Korea Institute of Machinery \& Materials, Daejeon, Korea \\
(\{rightcore, dyk, chkim78\}@kimm.re.kr)  {\small${}^{*}$ Corresponding author}}


\abstract{
    This paper presents a robotic assembly framework that combines Vision-Language Models (VLMs) with imitation learning for assembly manipulation tasks. Our system employs a gripper-equipped robot that moves in 3D space to perform assembly operations. The framework integrates visual perception, natural language understanding, and learned primitive skills to enable flexible and adaptive robotic manipulation. Experimental results demonstrate the effectiveness of our approach in assembly scenarios, achieving high success rates while maintaining interpretability through the structured primitive skill decomposition.
}



\maketitle


\section{Introduction}

Robotic assembly tasks represent one of the most challenging problems in robotics, requiring precise manipulation capabilities combined with sophisticated reasoning about complex multi-step processes. Unlike simple pick-and-place tasks, assembly tasks demand long-term planning that spans multiple sequential actions, where each step must be carefully coordinated with previous and subsequent operations. Furthermore, these tasks require physical understanding of component interactions and spatial relationships between parts \cite{beltran2020variable,thomas2018learning,luo2019reinforcement}.

Vision-Language Models (VLMs) have emerged as powerful tools that bridge visual perception and high-level reasoning, offering significant advantages for robotic applications. These models excel at processing visual information while understanding natural language instructions, making them well-suited for complex manipulation tasks. Recent work has successfully applied VLMs to various robotic manipulation scenarios, demonstrating their ability to understand scene context and generate appropriate action sequences \cite{brohan2023rt,li2023vision,gao2024physically,hu2023look,nasiriany2024pivot,liu2024moka}.

 \begin{figure}[t]    \noindent
    \includegraphics[width=\linewidth]{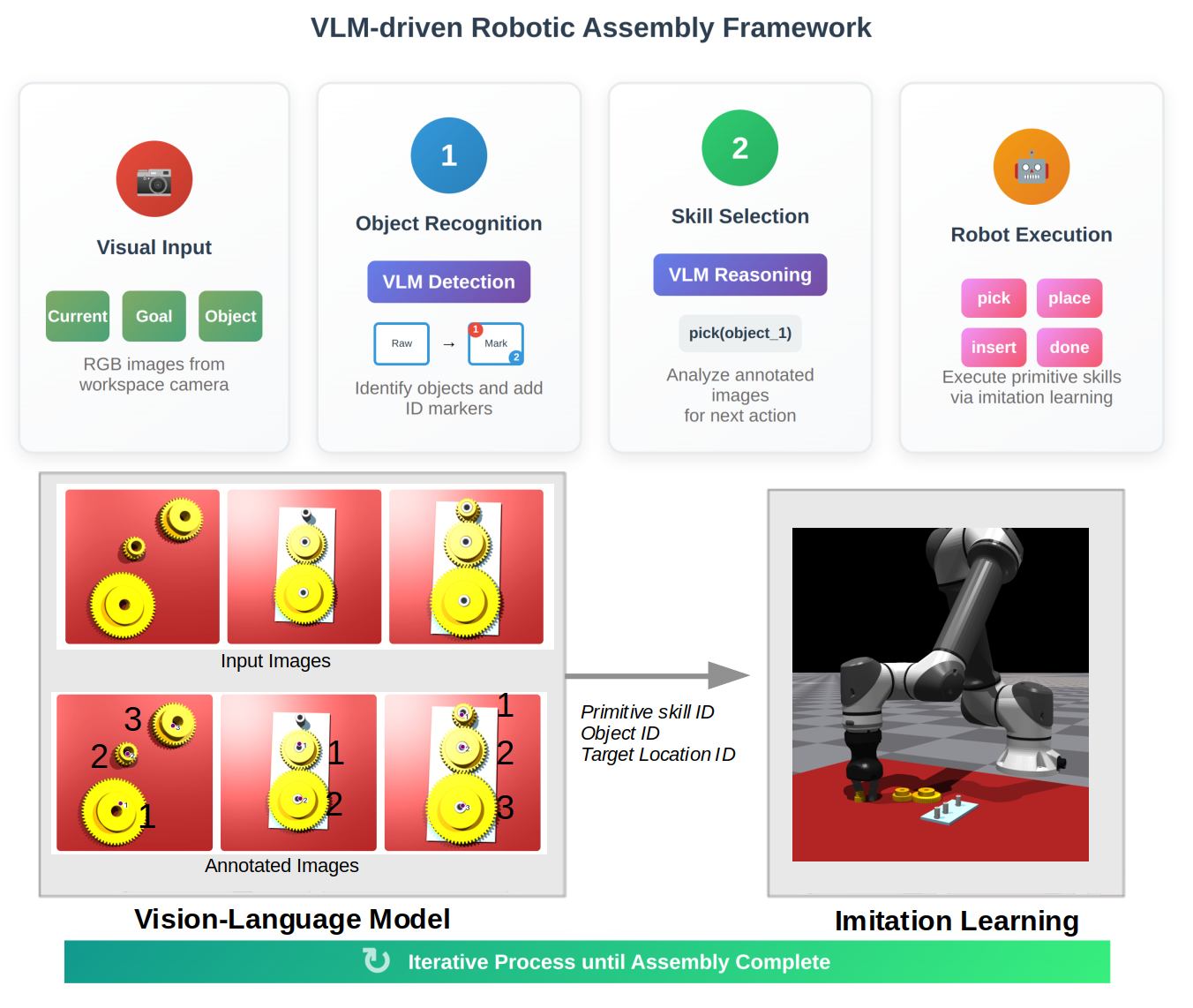}
    \caption{\label{framework_overview} VLM-driven robotic assembly framework showing the iterative process from visual input through two-stage VLM processing to skill execution.}
 \end{figure}
 The key strength of VLMs lies in their capacity for image-based state recognition, analyzing visual scenes to extract semantic information about object positions, orientations, and relationships. By incorporating contextual information such as previous state history, VLMs can perform sophisticated planning that accounts for immediate constraints and long-term objectives. However, existing approaches often treat skill selection as a monolithic process, limiting effectiveness in complex assembly scenarios requiring hierarchical decomposition and precise spatial reasoning.

 In this work, we propose a hierarchical VLM-driven framework that systematically addresses skill selection and parameterization for robotic assembly tasks. Our system leverages image observations and robot primitive actions to make informed decisions about skill execution and parameterization across long-horizon assembly sequences. Each skill is parameterized with specific spatial and temporal constraints, such as pick locations and target placement positions, enabling fine-grained control while maintaining adaptability.
 
 Our approach employs a two-stage VLM architecture that decomposes skill selection into specialized functions. The first stage performs visual scene analysis and object recognition with spatial marking. The second stage conducts skill reasoning and parameter selection based on annotated visual inputs, determining optimal skill sequences for assembly objectives. This hierarchical decomposition considers part dependencies, spatial constraints, and temporal relationships to generate coherent skill sequences.
 
 The integration of VLM-guided planning with imitation learning creates a robust framework combining abstract reasoning with precise skill execution. This approach handles high-level planning decisions about skill dependencies while ensuring reliable execution through learned primitive policies. The resulting framework addresses the complex challenges of long-horizon skill selection in robotic assembly domains.


\section{Methods}

\subsection{System Overview}

Our robotic assembly framework operates in a table-top environment using a single robot equipped with a gripper for object manipulation. The complete assembly process is illustrated in Figure \ref{framework_overview}, which shows the integration of visual perception, VLM processing, and skill execution components. The system employs a two-stage Vision-Language Model approach that first identifies and marks objects in the workspace, then performs reasoning for skill selection based on these visual annotations.
Based on the selected object and target information, the robot approaches the designated area to perform the task, and the imitation learning skill corresponding to the selected skill ID is executed to accomplish the assembly task.

\subsection{VLM-driven Skill Selection Framework}

Our robotic assembly system employs a two-stage Vision-Language Model (VLM) approach that separates object recognition from decision-making processes. This framework consists of two sequential VLM processing stages followed by skill execution, enabling robotic assembly operations through visual understanding and iterative decision-making.

\subsubsection{Stage 1: Object Recognition and Visual Annotation}

The first stage of our VLM processing focuses on identifying and marking task-relevant objects within the workspace. The VLM receives visual input and generates annotated images:

\begin{equation}
\textbf{I}_{\text{raw}} = \{I_{\text{object}},I_{\text{current}}, I_{\text{goal}}\}
\end{equation}

\noindent where $I_{\text{object}} \in \mathbb{R}^{H \times W \times \text{3}}$ , $I_{\text{current}} \in \mathbb{R}^{H \times W \times \text{3}}$ and $I_{\text{goal}} \in \mathbb{R}^{H \times W \times \text{3}}$ represent the task object, current assembly state and target state, respectively.

The object recognition VLM processes these images to identify task objects and generates point annotations:

\begin{equation}
\textbf{P} = \text{VLM}_{\text{recognition}}(I_{\text{object}},I_{\text{current}}, I_{\text{goal}}, P_{\text{obj}})
\end{equation}

\noindent where ${\textbf{P}} = \{\mathbf{p}_1, \mathbf{p}_2, \ldots, \mathbf{p}_n\}$ represents the set of 2D point coordinates ${\mathbf{p}}_i = [x_i, y_i] \in \mathbb{R}^\text{2}$ corresponding to identified objects, and $P_{\text{obj}}$ represents a list of object names used in the task.

These point coordinates are then used to generate visually annotated images with numerical markers:

\begin{equation}
I_{\text{object}}^{ann},I_{\text{current}}^{ann},I_{\text{goal}}^{ann} = \text{VisualMarking}(I_{\text{object}},I_{\text{current}}, I_{\text{goal}}, \textbf{P})
\end{equation}

\noindent where $I_{\text{object}}^{ann}$, $I_{\text{current}}^{ann}$, and $I_{\text{goal}}^{ann}$ contains the original image overlaid with numerical ID labels at the identified locations.

This approach draws inspiration from mark-based visual prompting techniques \cite{nasiriany2024pivot,liu2024moka}. In our proposed method, we implement a marking system that identifies and annotates three critical components: the target object, the current state, and the desired goal state.
The mark-based visual prompting framework provides a structured approach to visual understanding by explicitly highlighting relevant regions and states within the visual input. Building on this concept, we expand the standard marking approach by adding temporal and state-contextual information. Our enhanced marking system works in the following manner:

\begin{itemize}
    \item Target object marking: Visual identification and boundary delineation of the primary object of interest
    \item Current state marking: Annotation of the present configuration, position, or condition of the target object
    \item Goal state marking: Visual representation of the desired final state or target configuration 
\end{itemize}

This marking approach enables the system to maintain spatial and temporal awareness throughout the task execution process, facilitating more robust visual reasoning and decision-making capabilities.

For point annotation generation, we employ \cite{deitke2024molmo} in this work, although alternative methods can be readily substituted depending on the specific application requirements and constraints. The flexibility in annotation generation methods ensures broader applicability across different domains and use cases.

\subsubsection{Stage 2: Multi-Modal Input Processing and Skill Selection}




The second stage VLM receives enhanced multi-modal inputs, formally defined as:

\begin{equation}
\textbf{I}_\text{multi} = \{I_{\text{object}}^\text{ann},I_{\text{current}}^\text{ann},I_{\text{goal}}^\text{ann} , T_{\text{task}}\}
\end{equation}

\noindent where $T_{\text{task}}$ denotes the structured natural language prompt incorporating the marking information.

The VLM reasoning process generates the next primitive action and target identifications:
\begin{equation}
(skill, ID_{\text{obj}}, ID_{\text{target}}) = \text{VLM}_{\text{reason}}(\textbf{I}_{\text{multi}}, P_{\text{task}})
\end{equation}

\noindent where $skill \in \textbf{S}$, $ID_{\text{obj}}$ identifies the manipulation object, and $ID_{\text{target}}$ specifies the target location or object for placement operations.

The primitive skill set is defined as:
\begin{equation}
\textbf{S} = \{\text{pick}, \text{place}, \text{insert}, \text{done}, \text{init}\}
\end{equation}

Each primitive skill serves a specific function in the assembly process:
\begin{itemize}
    \item \textbf{pick}$(\text{object\_id})$: Grasps an object identified by its numerical marker
    \item \textbf{place}$(\text{target\_id})$: Places the currently held object at the target location specified by its numerical marker
    \item \textbf{insert}$(\text{target\_id})$: Performs insertion operations for assembly tasks at the marked target location
    \item \textbf{done}(): Signals completion of the current assembly task
    \item \textbf{init}(): Initializes the robot to a predefined starting configuration
\end{itemize}

In this work, we present a limited set of primitive actions including pick, place, insert, done, and init. However, this primitive set can be readily extended to incorporate additional actions specifically designed for complex assembly tasks, enabling broader applicability to diverse robotic manipulation scenarios.

\begin{figure*}
\begin{center}
\includegraphics[width=\textwidth]{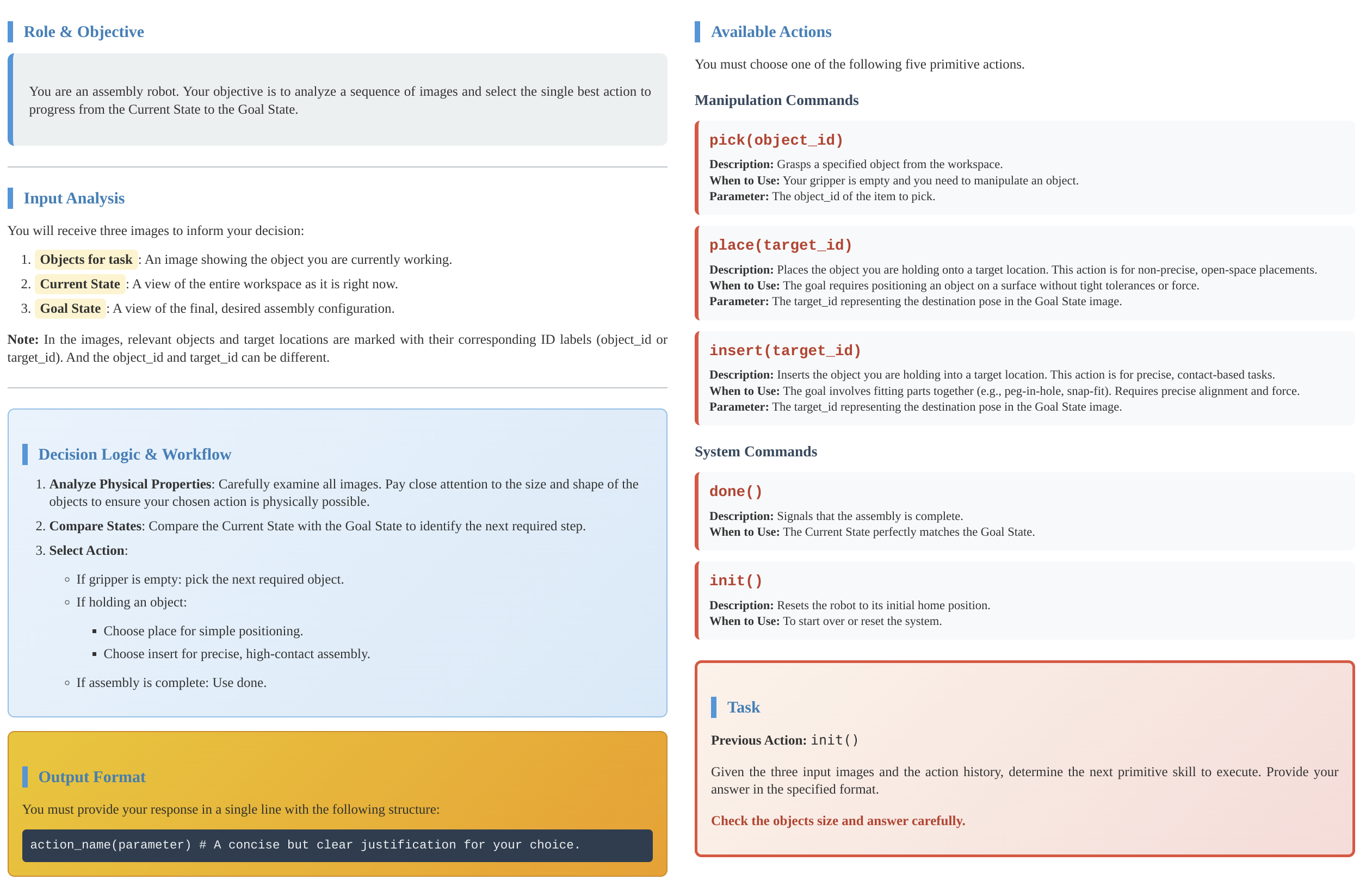}
\caption{\label{vlm_propt}Prompt architecture integrating task description, state analysis, and action specification }
\end{center}
\end{figure*}

The natural language prompt $P_{\text{task}}$ used in our system is specifically designed to leverage the visual annotations from Stage 1 and guide the VLM's reasoning process, as illustrated in Figure \ref{vlm_propt}. The prompt structure includes task description, current state analysis with reference to numerical markers, available actions with ID-based parameters, and output format specifications.

The prompt incorporates several key components: (1) task context and objectives, (2) current visual state description with reference to numerical markers, (3) available primitive skills with ID-based parameters, (4) decision logic workflow emphasizing the use of visual annotations, and (5) expected output format for skill selection and ID specification.

\subsection{Skill Execution via Imitation Learning}

Each primitive skill is executed using imitation learning models trained on human demonstrations. Imitation learning offers several significant advantages including the ability to generate task-specific skills that are precisely adapted to particular manipulation requirements. This approach has been successfully applied to various robotic tasks across different domains, demonstrating its versatility and effectiveness \cite{zhao2023learning,fu2024mobile,chi2023diffusion}.

The skill execution process is formulated as:
\begin{equation}
\boldsymbol{\tau} = \pi_{skill}(I^{\text{cam}}_{t}, \text{s}_{t})
\end{equation}

\noindent where:
\begin{itemize}
    \item $\boldsymbol{\tau} = \{\mathbf{u}_{t}, \mathbf{u}_{t+1}, \ldots, \mathbf{u}_{t+T}\}$ is the robot trajectory
    \item $\mathbf{u}_i \in \mathbb{R}^\text{n}$ represents the robot command at time step $i$
    \item $\pi_{skill}$ is the policy network for skill $s$
    \item $I^{\text{cam}}_{t}$ denotes the camera image input
    \item $\text{s}_{t}$ represents the current robot state
\end{itemize}

Our implementation utilizes action chunking \cite{zhao2023learning} for learning, where the policy predicts sequences of actions rather than single actions at each time step. This approach improves temporal consistency and reduces compounding errors. 





\subsection{Iterative Assembly Process with VLM}

The complete framework operates through an iterative manner that continues until task completion. At each iteration, the system first processes the current images through Stage 1 VLM for object recognition and marking, then analyzes the annotated images through Stage 2 VLM to determine the next required skill and its necessary parameters. The selected skill is then executed using the corresponding imitation learning model, and the process repeats until the assembly task is finished.


By utilizing the ID information of the target object and target location inferred by the VLM, approximate values for the respective target positions can be obtained. Before executing the IL model, the system moves to the corresponding location with a z-axis offset, then executes the IL model. Through this process, the VLM estimates the state from a high level and establishes an overall task plan, while positioning the robot to a local location where the IL model can operate effectively, thereby providing a state conducive to IL model execution. This approach combines the inference results of the VLM with the execution process of the IL model, enabling the robot to perform tasks accurately.

The iterative process can be described as:
\begin{algorithm}
    \caption{VLM-driven Robotic Assembly}
    \begin{algorithmic}
    \State $skill = \text{init}()$
    \While{$skill \neq \text{done}()$}
    \State $\textbf{P} = \text{VLM}_{\text{recognition}}(I_{\text{object}}, I_{\text{current}}, I_{\text{goal}}, P_{\text{obj}})$
    \State $\textbf{I}_{\text{annotated}} = \text{VisualMarking}(I_{\text{object}}, I_{\text{current}}, I_{\text{goal}}, \textbf{P})$
    \State $(skill, ID_{\text{obj}}, ID_{\text{target}}) $
    \State \hspace{2.5cm} $ =\text{VLM}_{\text{reason}}(\textbf{I}_{\text{annotated}}, P_{\text{task}})$
    
    \If{$skill = \text{pick}$}
        \State $Move(ID_{\text{obj}})$
        \State $\boldsymbol{\tau} = \pi_{s}(I^{\text{cam}}_{t}, \text{s}_{t})$
    \ElsIf{$skill = \text{place}$ or $skill = \text{insert}$}
        \State $Move(ID_{\text{obj}}, ID_{\text{target}})$
        \State $\boldsymbol{\tau} = \pi_{s}(I^{\text{cam}}_{t}, \text{s}_{t})$
    \EndIf

    
    \EndWhile
    \end{algorithmic}
\end{algorithm}




\section{Experiments}

To validate the proposed method, we conducted two distinct experiments in both simulation and real-world environments. The first experiment focuses on VLM-based primitive selection for gear assembly tasks, while the second experiment evaluates the integration capability with Imitation Learning (IL) for complete task execution.

\subsection{VLM-based Primitive Selection}

This experiment evaluates the effectiveness of VLM-based primitive selection for gear assembly tasks. We utilized GPT-4.1-2025-04-14 model and GPT-5-mini-2025-08-07 model as the vision-language model and conducted evaluations across one simulation environment and two real-world environments.

\subsubsection{Experimental Setup}

The evaluation criteria were binary: success if the appropriate primitive was selected, failure otherwise. Each experimental condition was assessed through two sequential stages:
\begin{itemize}
\item \textbf{Pick task}: Evaluation of whether the system selects the appropriate object for grasping
\item \textbf{Insert task}: Evaluation of whether the system identifies the correct insertion location
\end{itemize}

\subsubsection{Results}

\begin{figure}[htb]
    \begin{center}
    \includegraphics[width=7.5cm]{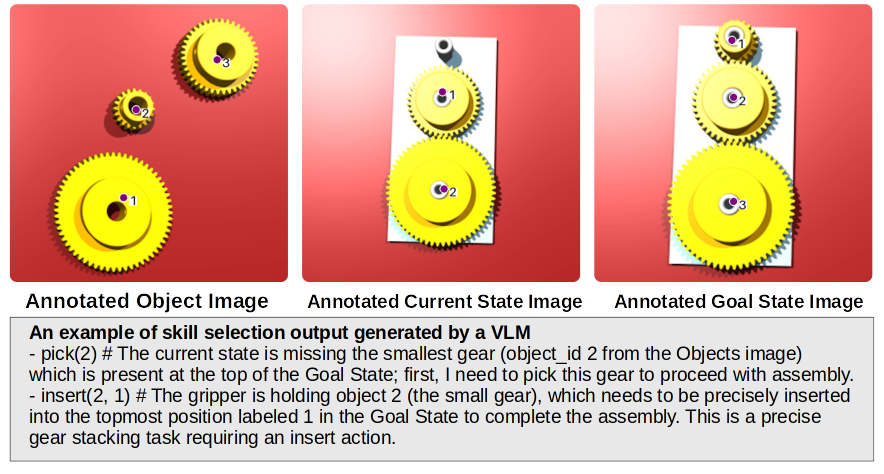}
    \caption{\label{sim_ex1}VLM-based primitive selection results for simulation environment.}
    \end{center}
\end{figure}

\begin{figure}[htb]
    \begin{center}
    \includegraphics[width=7.5cm]{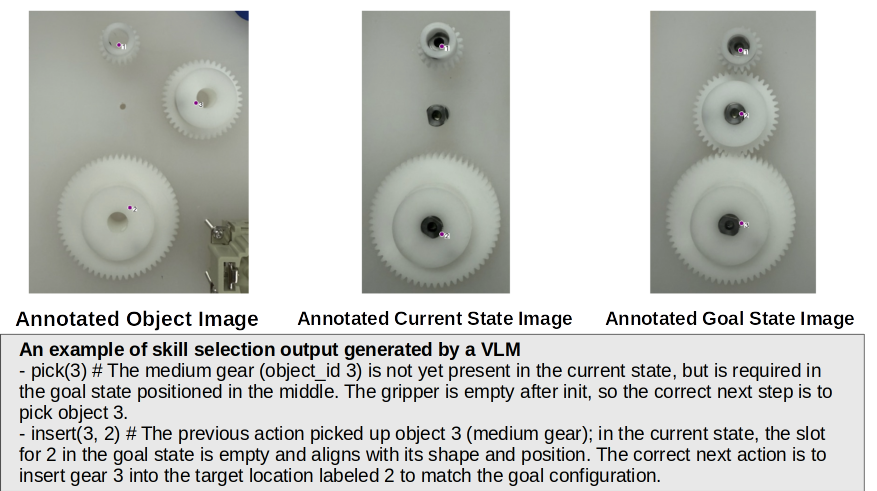}
    \caption{\label{real_ex1}VLM-based primitive selection results for real environment 1.}
    \end{center}
\end{figure}

\begin{figure}[htb]
    \begin{center}
    \includegraphics[width=7.5cm]{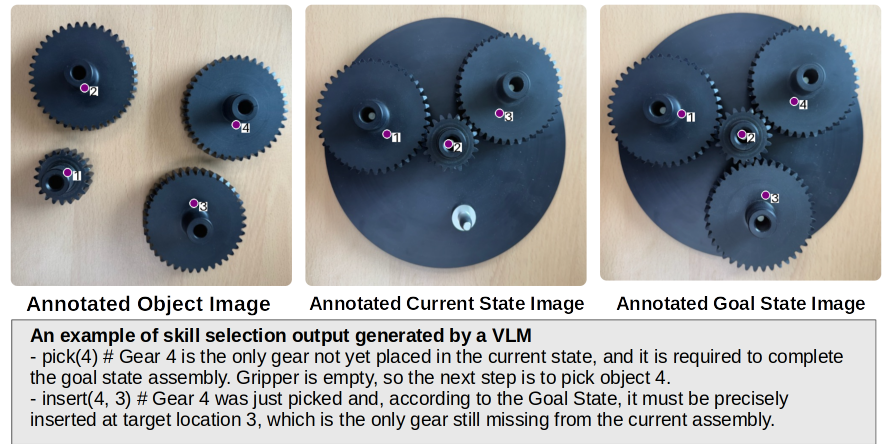}
    \caption{\label{real_ex2}VLM-based primitive selection results for real environment 2.}
    \end{center}
\end{figure}

The experimental results across different environments are summarized in Table \ref{tab:primitive_selection} and examples of skill selection output generated by a VLM are shown in Figure \ref{sim_ex1}, Figure \ref{real_ex1}, and Figure \ref{real_ex2}.

\begin{table}[htb]
    \setlength{\extrarowheight}{0.75ex}
    \caption{VLM-based primitive selection success rates across different environments.}
    \label{tab:primitive_selection}
    \begin{center}
    \begin{tabu}to\linewidth{|X[c]|X[c]|X[c]|X[c]|X[c]|}\hline
    \multirow{2}{*}{Env.} & \multicolumn{2}{c|}{GPT-4.1} & \multicolumn{2}{c|}{GPT-5-mini} \\\cline{2-5}
     & Pick & Insert & Pick & Insert \\\hline
    Sim. & 3/10 (30\%) & 3/10 (30\%) & 3/10 (30\%) & 10/10 (100\%) \\\hline
    Real 1 & 10/10 (100\%) & 3/10 (30\%) & 10/10 (100\%) & 10/10 (100\%) \\\hline
    Real 2 & 10/10 (100\%) & 0/10 (0\%) & 8/10 (80\%) & 10/10 (100\%) \\\hline
    \end{tabu}
    \end{center}
\end{table}


The results demonstrate significant differences in VLM performance between models and tasks. GPT-4.1 shows consistent high success rates for pick tasks in real-world environments (100\%), but struggles considerably with insertion tasks, achieving only 30\% success in Real-world 1 and 0\% in Real-world 2. In contrast, GPT-5-mini exhibits substantially improved performance across both task types, achieving perfect success rates (100\%) for insertion tasks in all environments and maintaining strong pick task performance (80-100\% in real-world settings).

Notably, both models show similar limitations in simulation environments for pick tasks (30\% success rate), suggesting inherent challenges in the simulated domain. However, GPT-5-mini's perfect insertion task performance in simulation (100\% vs GPT-4.1's 30\%) indicates significant advances in spatial reasoning capabilities. The dramatic improvement from GPT-4.1 to GPT-5-mini, particularly for insertion tasks, suggests that recent VLM developments have substantially enhanced precise spatial reasoning and manipulation planning capabilities. While domain transfer from simulation to real-world remains challenging for pick tasks, GPT-5-mini demonstrates that insertion task precision, previously a major limitation, can be effectively addressed with advanced VLM architectures.



\subsection{Integration with Imitation Learning}

This experiment validates the proposed framework's effectiveness under the assumption that VLM-guided primitive selection generates appropriate plans. Given well-reasoned skill sequences, object selections, and target locations from the VLM, we demonstrate that the integrated framework can successfully execute complete assembly tasks through learned primitive policies. The experiment aims to confirm the framework's suitability for hierarchical task decomposition by showing reliable primitive execution when provided with accurate high-level planning.

\subsubsection{Experimental Setup}


The IL policy was implemented using Diffusion Policy \cite{chi2023diffusion} with the following configuration:
\begin{itemize}
\item \textbf{Training data}: 10 demonstration trajectories per primitive task
\item \textbf{Training iterations}: 500 optimization steps
\item \textbf{Network architecture}: The noise prediction network employs a Diffusion Transformer (DiT) \cite{peebles2023scalable} architecture and isual features are extracted using the DINO feature extractor \cite{simeoni2025dinov3} to provide robust visual representations. The network reconstructs target trajectories by generating sequences of 16 consecutive actions.
\item \textbf{Input modalities}: Multi-modal sensory inputs comprising manipulator-mounted camera images, side-view and front-view camera observations, end-effector poses, and force measurements
\end{itemize}

The system receives VLM-generated skill IDs, object IDs, and location IDs to guide sequential task execution. Using the object IDs and location IDs, the robot navigates to the vicinity of the corresponding target points, and the imitation learning model associated with the specified skill ID is then executed.

\begin{figure}[t]
\begin{center}
\includegraphics[width=5.5cm]{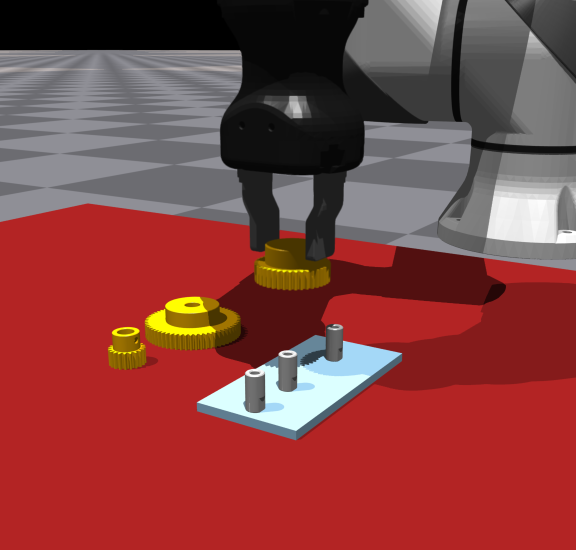}
\caption{\label{il_integration}Robotic assembly task execution in simulation environment using the proposed framework}
\end{center}
\end{figure}

Figure \ref{il_integration} illustrates the robot performing assembly tasks in simulation using the proposed method. While the baseline method (complete IL-based execution) failed to successfully complete gear assembly tasks, the proposed method achieved partial success in gear assembly operations.  The proposed VLM-guided primitive selection framework provides  advantages over monolithic IL approaches for complex manipulation tasks. The decomposition of complex tasks into manageable primitives, guided by visual scene understanding, enables more effective learning and execution of manipulation skills.






\subsection{Discussion}

The experimental results reveal several key insights across different VLM architectures:

\begin{enumerate}
\item \textbf{Object identification proficiency}: Both VLMs demonstrate robust object identification capabilities in real-world environments, confirming effective visual understanding and scene comprehension.

\item \textbf{Spatial reasoning advancement}: Significant performance variation between VLM architectures in precision tasks indicates that recent developments have substantially improved high-precision spatial reasoning capabilities previously limiting manipulation systems.

\item \textbf{Validated integration framework}: The integration with imitation learning consistently outperforms monolithic approaches, validating the effectiveness of hierarchical task decomposition through VLM-guided primitive selection regardless of the underlying model architecture.

\item \textbf{Effective sim-to-real transfer}: Advanced VLM architectures demonstrate minimal performance gaps between simulation and real-world environments, particularly in precision tasks, indicating that robust visual reasoning capabilities can effectively bridge domain differences and enable reliable sim-to-real transfer.
\end{enumerate}

These findings demonstrate the flexibility and scalability of our proposed framework. The performance variation across different VLM architectures validates that our hierarchical decomposition approach effectively leverages underlying capabilities without architectural constraints. This modular design enables seamless integration of advancing VLM technologies, allowing the framework to naturally benefit from future developments in visual reasoning. Future work should focus on enhancing simulation environment fidelity and exploring additional primitive skills.





\section{Conclusion}
This paper presented a hierarchical VLM-integrated robotic assembly framework that effectively combines visual perception, spatial reasoning, and learned primitive skills through structured task decomposition. The experimental validation demonstrates the framework's adaptability and scalability across different VLM architectures, confirming its capacity to leverage advancing foundation model capabilities without structural modifications.

The key contributions include: (1) a modular architecture that seamlessly integrates evolving VLM capabilities, (2) validated effectiveness of hierarchical primitive decomposition for manipulation tasks, and (3) interpretable decision-making through explicit skill selection and visual annotation processes. 
The framework establishes a foundation for VLM-driven manipulation systems that can evolve with advancing visual reasoning capabilities while maintaining reliable performance in diverse deployment scenarios. Future work will explore expanded primitive skill sets and validation in diverse industrial assembly contexts.

\section*{ACKNOWLEDGEMENT}
This work was supported by the Ministry of Trade, Industry and Energy (MOTIE) in 2025 under Grant RS-2024-00416440 for the project titled “Development of the AI-based autonomous task planning and robot teaching solution for highly complex manufacturing assembly process,” and by the National Research Council of Science and Technology under Grant NK254G for the project titled “Development of core technologies for robot general purpose task artificial intelligence (RoGeTA) framework.”

\bibliographystyle{unsrt}
\bibliography{ref.bib}

%




%

\end{document}